\documentclass[10pt,twocolumn,letterpaper]{article}

\usepackage{cvpr}              

\usepackage{textcomp}
\usepackage{stfloats}
\usepackage{url}
\usepackage{verbatim}
\usepackage{graphicx}
\usepackage{algorithm}
\usepackage{multicol}
\usepackage{multirow}
\usepackage{colortbl}
\usepackage{tcolorbox}
\usepackage{bm} 
\usepackage{hyperref}
\usepackage{booktabs}
\usepackage{enumitem}
\usepackage{balance}
\definecolor{cvprblue}{rgb}{0.21,0.49,0.74}

\def\paperID{7095} 
\def\confName{CVPR}
\def\confYear{2026}

\title{PhyVLLM: Physics-Guided Video Language Model \\ with Motion–Appearance Disentanglement
}

\author{
Yu-Wei Zhan\textsuperscript{1},
Xin Wang\textsuperscript{1}\thanks{Corresponding author},
Hong Chen\textsuperscript{1},
Tongtong Feng\textsuperscript{1},
Wei Feng\textsuperscript{1},
Ren Wang\textsuperscript{1}, \\ 
Guangyao Li\textsuperscript{1},
Qing Li\textsuperscript{2},
Wenwu Zhu\textsuperscript{1}\footnotemark[1] \\[2mm]
\textsuperscript{1}Department of Computer Science and Technology, Tsinghua University \\
\textsuperscript{2}Department of Electronic Engineering, Tsinghua University \\[1mm]
}

\begin{document}
\maketitle
\begin{abstract}
Video Large Language Models (Video LLMs) have shown impressive performance across a wide range of video-language tasks. However, they often fail in scenarios requiring a deeper understanding of physical dynamics. This limitation primarily arises from their reliance on appearance-based matching. Incorporating physical motion modeling is crucial for deeper video understanding, but presents three key challenges: (1) motion signals are often entangled with appearance variations, making it difficult to extract clean physical cues; (2) effective motion modeling requires not only continuous-time motion representations but also capturing physical dynamics; and (3) collecting accurate annotations for physical attributes is costly and often impractical.
To address these issues, we propose PhyVLLM, a physical-guided video-language framework that explicitly incorporates physical motion into Video LLMs. Specifically, PhyVLLM disentangles visual appearance and object motion through a dual-branch encoder. To model
physical dynamics over time, we incorporate a Neural Ordinary Differential Equation (Neural ODE) module, which generates differentiable physical dynamic representations. The resulting motion-aware representations are projected into the token space of a pretrained LLM, enabling physics reasoning without compromising the model’s original multimodal capabilities.
To circumvent the need for explicit physical labels, PhyVLLM employs a self-supervised manner to model the continuous evolution of object motion.
Experimental results demonstrate that PhyVLLM significantly outperforms state-of-the-art Video LLMs on both physical reasoning and general video understanding tasks, highlighting the advantages of incorporating explicit physical modeling.
\end{abstract}    
\section{Introduction}
\label{sec:intro}

\begin{figure}[t]
\centering
\begin{minipage}{0.98\linewidth}\centering
\centerline{\includegraphics[height=9.5cm]{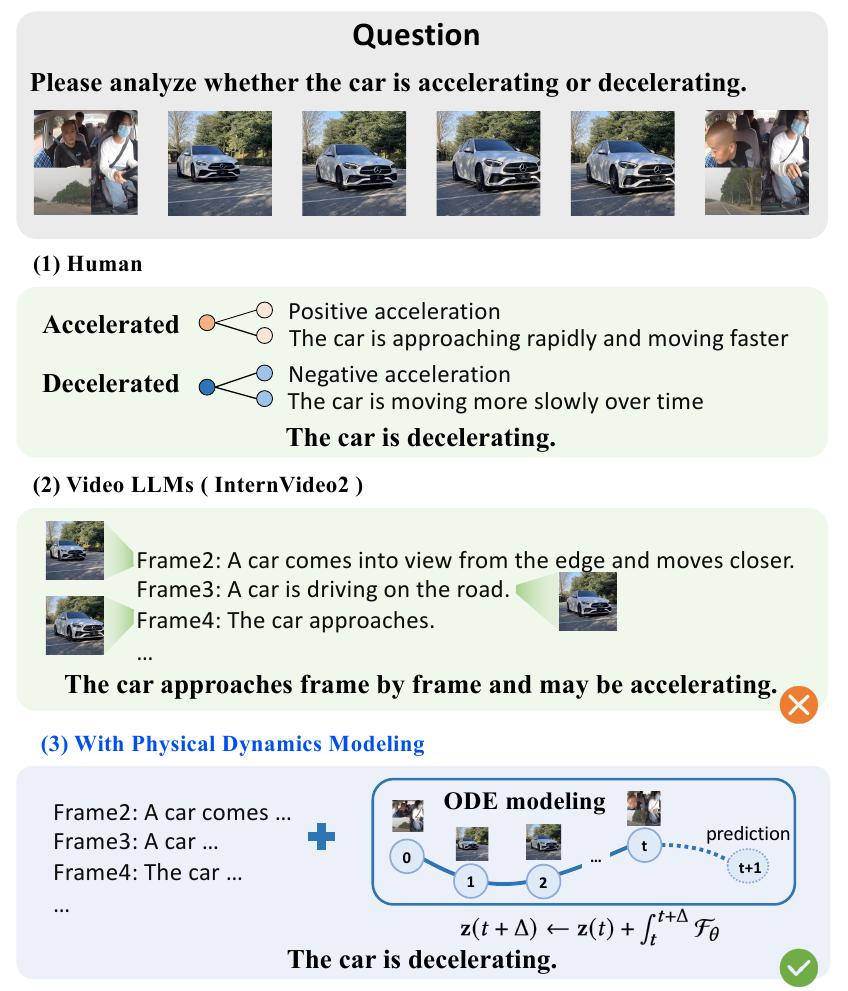}}
\end{minipage}
\vspace{-0.2cm} 
\caption{Example of acceleration vs. deceleration recognition. (a) Humans determine whether an object is accelerating or decelerating by reasoning over physical attributes, such as the sign of acceleration. (b) A Video LLM fails to distinguish the two cases due to a lack of motion modeling. (c) Our method explicitly models dynamic motion using a Neural ODE and successfully infers the correct physical state.
}\label{intro}
\vspace{-0.5cm} \medskip\end{figure}

Video Large Language Models (Video LLMs) have recently achieved impressive results across various multimodal tasks, including video question answering, captioning, and retrieval \cite{li2023mimic,ko2023large,maaz2023video}. These results highlight their strong capability in capturing appearance semantics and aligning visual content with natural language.

Most existing video LLMs adopt a frame-based encoding strategy, where videos are processed as sequences of static images using pretrained visual backbones, and then aligned with natural language instructions via instruction tuning. This design effectively leverages large-scale video-text data to learn semantic correlations, achieving significant success on appearance-driven tasks. To strengthen temporal modeling, some methods further incorporate optical flow \cite{jin2024video}, which provides displacement and directional cues between adjacent frames. However, optical flow is inherently short-term, pixel-level motion estimation. It struggles to capture long-term dynamics and higher-order motion patterns, such as acceleration, and lacks the ability to predict future states.
This limitation indicates that, although current Video LLMs exhibit strong performance on multimodal tasks, they still struggle in scenarios requiring dynamics modeling. As 
illustrated in Fig.~\ref{intro} (b), the video clip shows a car in a real-world environment that is actually \emph{decelerating}. However, the InternVideo2 \cite{wang2024internvideo2}  predicts that the car is \emph{accelerating}. This occurs because “acceleration” and “deceleration” exhibit strong visual similarity, both manifested as the car moving in the same direction. In other words, frame-by-frame visual feature representations overlap significantly, leading to misclassification. In contrast, humans can easily reason about the sign of acceleration to distinguish whether a vehicle is speeding up or slowing down. This comparison further underscores the limitations of current Video LLMs in dynamic modeling.

Physics-based dynamic modeling plays a crucial role in enhancing video understanding by providing detailed motion cues. Such modeling can capture variations in velocity, directions of acceleration, and trajectory continuity, thereby revealing latent physical laws that are difficult to observe directly from appearance. As shown in Fig.~\ref{intro} (c), when analyzing a video of a car approaching a crosswalk, models relying solely on frame-by-frame appearance may fail to determine whether the car is accelerating through or decelerating to stop. In contrast, physics-based dynamic modeling allows us to infer a negative acceleration from the latent dynamical function $F_{\theta}$, thereby concluding that the car is decelerating. Leveraging such modeling enables Video LLMs not only to distinguish between visually similar but dynamically distinct motion states, but also to improve their understanding of complex behaviors in real-world scenarios. For example, a passenger is leaning forward at the end of the video; this can be attributed to the inertial effect of deceleration, revealing a causal link between human behavior and vehicle dynamics. This physically guided explanatory ability improves robustness on physics-aware tasks and moves beyond shallow recognition that relies on visual–language alignment.

Therefore, to enable deeper physical reasoning within video LLMs, explicit modeling of physical dynamics is essential. However, integrating physical dynamics into Video LLMs introduces some key challenges:
(1) Modeling physical dynamics requires accurately capturing object motion. However, motion signals in videos are often entangled with visual appearance variations, such as changes in texture, lighting, or viewpoint. This makes it difficult to extract motion as a clean and reliable cue. 
(2) Motion representation remains a core challenge. Most current Video LLMs are built on Transformer that extracts patch-based features from static frames. This design works well for appearance and semantics, but fails to build continuous-time motion representations or capture physical dynamics. A key challenge is to design representations that preserve temporal coherence while modeling dynamic processes.
(3) Lack of physical supervision. Collecting accurate annotations for physical attributes (e.g., acceleration, contact forces) is costly and often impractical.  This data sparsity poses a major barrier to learning physically grounded representations and limits the model’s ability to generalize.

To address these challenges, we introduce PhyVLLM, a physical-guided video-language framework that explicitly decouples physical motion from visual appearance and models dynamic motion in a continuous and differentiable manner. The framework begins with a dual-branch encoder, where one branch extracts static appearance features and the other captures dynamic motion cues. To model physical dynamics over time, we incorporate a Neural Ordinary Differential Equation (Neural ODE) module, which simulates object trajectories as continuous-time processes and generates differentiable physical representations. These representations are projected into the token space of a frozen LLM and fused via a lightweight LoRA module, enabling efficient fine-tuning and seamless integration with downstream video-language tasks. This design allows the model to incorporate physical priors while retaining its general multimodal reasoning capabilities. 
To address the lack of annotated physical labels in existing datasets, we employed a self-supervised physical consistency loss during the training of PhyVLLM. 
We evaluate PhyVLLM against several state-of-the-art Video LLMs on the simulation benchmark PhyBench, as well as on several general video understanding benchmarks. Results show that our method significantly enhances physical reasoning capabilities.

Our contributions are summarized as follows:
\begin{itemize}
    \item We propose a physical-guided video understanding framework that explicitly models physical dynamics, enabling Video LLMs to reason about dynamic motions.
    \item We disentangle motion dynamics from visual appearance and leverage Neural ODEs to construct continuous-time motion representations, serving as physical priors within the tokenized multimodal space.
    \item We introduce a self-supervised learning paradigm that aligns predicted motion trajectories with underlying physical dynamics through physical-consistency constraints, eliminating the need for manual annotations.
    \item We demonstrate the effectiveness of our method through extensive evaluations on PhyBench and general video understanding benchmarks.
\end{itemize}

\section{Related Work}

\subsection{Video Large Language Models}

To harness the full potential of LLMs for video understanding, researchers have proposed a variety of strategies. Current efforts largely fall into two main categories: \textit{LLM-based intelligent video agents} and \textit{instruction-tuned video-language models}.

\textbf{LLM-based Intelligent Video Agents} leverage the powerful sequence modeling and contextual understanding capabilities of LLMs to reconstruct the architecture and processes of video understanding from a novel perspective~\cite{tang2025hawk, park2024too, wang2023chatvideo}. For instance, HuggingGPT~\cite{shen2023hugginggpt} employs ChatGPT~\cite{openai2022chatgpt} for task planning and decomposition, dynamically selecting specialized models from Hugging Face based on function descriptions. This architecture fully utilizes ChatGPT’s strengths in language comprehension, reasoning, and interaction, along with Hugging Face’s extensive AI model ecosystem. Video ChatCaptioner~\cite{chen2023video} creates an interactive dialogue system between ChatGPT and BLIP-2~\cite{li2023blip}, generating rich video descriptions. In this setup, ChatGPT acts as a controller, selecting key frames from videos and posing relevant questions, while BLIP-2 serves as the visual comprehension module to answer them.

\begin{figure*}[t]
\centering
\begin{minipage}{1.0\linewidth}\centering
\centerline{\includegraphics[height=7cm]{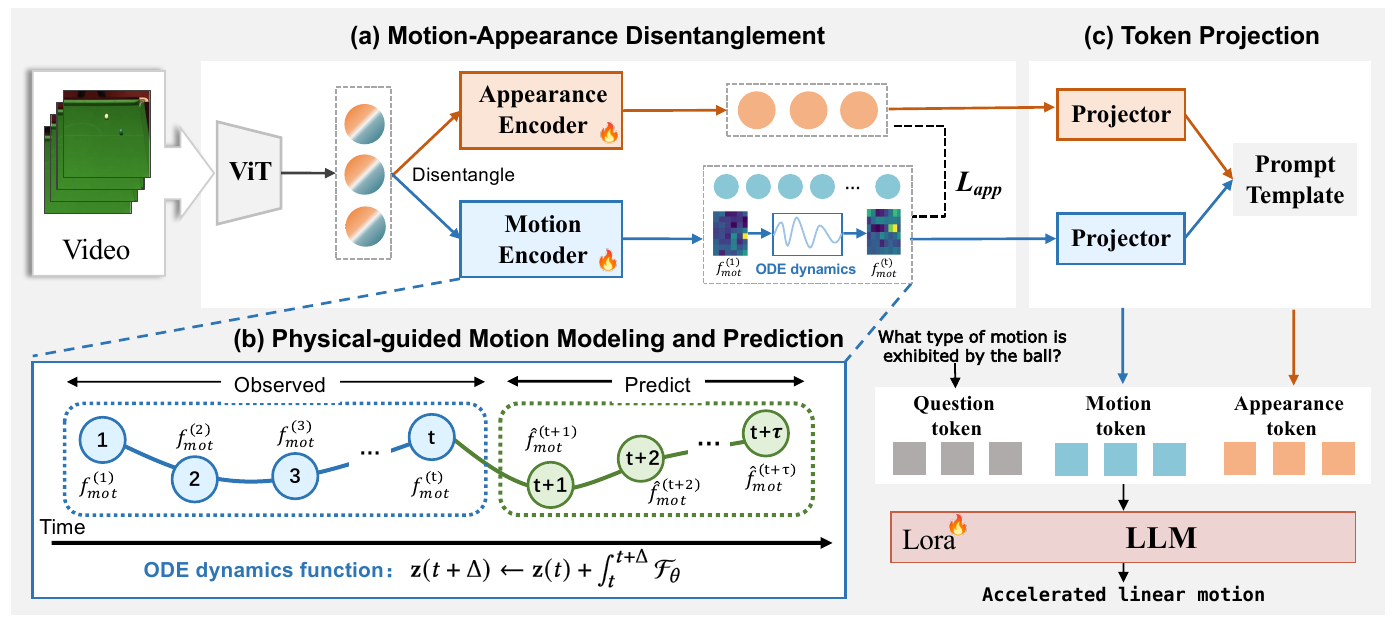}}
\end{minipage}
\vspace{-0.2cm} 
\caption{Overview of the proposed PhyVLLM framework. It consists of three main components: (a) Motion-Appearance Disentanglement, which separates appearance cues and dynamic motion patterns via a dual-branch encoder; (b) Physical-Guided Motion Modeling and Prediction, where a Neural ODE module continuously models object dynamics from the motion features; and (c) Token Projection, which maps the disentangled features into the token space of a pretrained LLM using lightweight adapters, enabling seamless integration while preserving compatibility with the frozen backbone.}\vspace{-0.2cm} \label{framework}\medskip\end{figure*}

\textbf{Instruction-Tuned Video-Language Models} are designed to bridge visual encoders and LLMs, enabling comprehensive video understanding through fine-tuning on large-scale, high-quality instruction datasets~\cite{li2024llama, huang2024vtimellm, li2023mimic, ko2023large, maaz2023video}. These methods often incorporate boundary-aware training strategies to enhance video comprehension~\cite{lyu2023macaw, zhang2023video, huang2024vtimellm, chen2023videollm, jin2024chat, sun2023fine}. 
mPLUG-video~\cite{xu2023youku} is trained and fine-tuned on Youku-mPLUG, the largest high-quality Chinese video-language dataset to date. Its modular decoder-only architecture, combined with instruction tuning, significantly enhances video understanding capabilities. Video-ChatGPT~\cite{maaz2023video} integrates video-adapted visual encoders with LLMs to form a multimodal dialogue system capable of generating detailed video-based conversations.


However, current Video LLMs struggle with physical reasoning. They mainly rely on pattern matching over visual and semantic features, without modeling physical dynamics, leading to failures in tasks like motion prediction and distinguishing physical behaviors.

\subsection{Physical-Guided Video Understanding}

Physical laws govern the motion, interaction, and transformation of objects in the real world and are thus essential for deep video understanding. For example, Hofherr et al.~\cite{hofherr2023neural} propose neural implicit representations for modeling appearances and inferring physical parameters from dynamic planar scenes. Unlike data-hungry approaches, their method achieves high efficiency by estimating physical properties from only a single video. Le Guen et al. introduce PhyDNet~\cite{guen2020disentangling}, a model that incorporates a recurrent physical cell inspired by data assimilation techniques. By enforcing partial differential equation constraints during prediction, PhyDNet generalizes physical reasoning beyond pixel-level observations. Aoyagi et al.~\cite{aoyagi2021spatio} 
adopts a Mixture-of-Experts framework with multiple spatiotemporal expert branches, using pixel-level and expert-level attention to adaptively integrate outputs from different branches based on the underlying physics.

However, existing research on integrating physical knowledge into video modeling remains limited to small-scale models. \textit{No current large video LLMs successfully incorporate explicit physical constraints.}
While these small-scale models demonstrate impressive performance on specific tasks, they face challenges in generalization and complex scene understanding. In contrast, large video LLMs excel in representation and generalization but lack an explicit understanding of physical dynamics, leading to poor performance on dynamic motion reasoning tasks.


\section{PhyVLLM}
\subsection{Overview}

We propose PhyVLLM, a physical-guided framework that disentangles dynamic motion from visual appearance and models dynamic motion in a continuous and differentiable manner. Specifically, our framework consists of three key components, as illustrated in Figure \ref{framework}. (a) Motion-Appearance Disentanglement. We adopt a dual-branch encoding strategy to disentangle static appearance information and dynamic motion features from the input video. (b) Physical-Guided Motion Modeling and Prediction. To explicitly model the temporal evolution of motion in a continuous and physically guided manner, we introduce a Neural ODE module. 
(c) Physics-aware Tokenization.
The learned motion and appearance features are then projected into the token space of a pretrained LLM. 
Further details of each module are provided in the subsequent sections.

\subsection{Motion-Appearance Disentanglement}


Physical motion plays a fundamental role in video understanding tasks. Unlike static images, videos inherently capture temporal evolution, encoding how objects move or interact over time~\cite{guen2020disentangling}. This dynamic information reflects the underlying physical laws governing the scene and offers critical guidance for downstream tasks. However, motion features are often entangled with appearance information such as object textures, lighting conditions, and background clutter, making it difficult for video LLMs to exploit the physical structure embedded in the temporal dynamics. To address this challenge, we propose a disentangled representation strategy that explicitly separates motion-related and appearance-related information within videos. Specifically, we design two parallel encoding branches: a motion encoder and an appearance encoder. Both encoders operate on the same video input but are specialized for different roles. The motion encoder captures temporally varying patterns that reflect physical dynamics, while the appearance encoder focuses on temporally stable visual attributes.

\paragraph{Architecture Design.}
Both the appearance encoder $E_{\text{app}}$ and the motion encoder $E_{\text{mot}}$ are built upon a shared visual backbone to ensure consistent low-level feature extraction and parameter efficiency. Specifically, we adopt a pretrained Vision Transformer (ViT) as the common feature extractor, which processes each video frame independently into patch-level embeddings. On top of this shared backbone, we attach two lightweight task-specific heads to model static and dynamic aspects, respectively. The appearance encoder $E_{\text{app}}$ is responsible for capturing the static content of the video, which remains largely consistent across frames. In our implementation, we use a shallow MLP over frame-level features to extract stable descriptors. 
In contrast, the motion encoder $E_{\text{mot}}$ operates on the full sequence to extract features that describe how objects move over time. This encoder is designed to be temporally sensitive and aware of frame-to-frame transitions. 
We implement $E_{\text{mot}}$ by stacking additional transformer blocks with temporal attention modules to capture inter-frame dependencies.

Formally, let $ViT_{\text{base}}$ denote the shared transformer encoder. Then, for each frame $I_t$, we compute patch-level features $\mathbf{h}^{(t)} = ViT_{\text{base}}(I_t)$. The two branches proceed as:
\begin{align}
f_{\text{app}}^{(t)} &= E_{\text{app}}(\mathbf{h}^{(t)}) \\
\{f_{\text{mot}}^{(t)}\}_{t=1}^T &= E_{\text{mot}}(\{\mathbf{h}^{(t)}\}_{t=1}^T)
\end{align}

This design preserves architectural simplicity while promoting functional disentanglement between static appearance encoding and dynamic motion modeling. The resulting motion sequence $\{f_{\text{mot}}^{(t)}\}_{t=1}^T\}$ serves as input to the subsequent Neural ODE-based physical modeling module described in the following section.

\paragraph{Disentanglement Loss.}
To disentangle motion from appearance, we adopt the Hilbert--Schmidt Independence Criterion (HSIC) \cite{gretton2005measuring}. HSIC serves as a practical proxy for minimizing mutual information between motion features $F_{\text{mot}}$ and appearance features $F_{\text{app}}$. Given a minibatch $\{(f^{\text{mot}}_i, f^{\text{app}}_i)\}_{i=1}^n$, we compute Gram matrices $K_{ij}=k(f^{\text{mot}}_i,f^{\text{mot}}_j)$ and $L_{ij}=l(f^{\text{app}}_i,f^{\text{app}}_j)$ with RBF kernels, and apply the centering matrix $H = I_n - \tfrac{1}{n}\mathbf{1}\mathbf{1}^\top$. The empirical estimate is
\begin{equation}
\mathrm{HSIC}(F_{\text{mot}},F_{\text{app}})
= \frac{1}{(n-1)^2}\,\mathrm{tr}(KHLH).
\end{equation}
We define the appearance disentanglement loss as
\begin{equation}
\mathcal{L}_{\text{app}}=\mathrm{HSIC}(F_{\text{mot}},F_{\text{app}}),
\label{app_loss}
\end{equation}
which enforces independence between motion and appearance representations.

\subsection{Physical-guided Motion Modeling and Prediction}
\label{sec:neural_ode}

\paragraph{Preliminaries of Neural ODEs.}
Neural ODEs~\cite{chen2018neural} provide a principled framework for modeling continuous-time processes using data-driven dynamics. Unlike traditional discrete models (e.g., RNNs, Transformers), which process sequences step-by-step, Neural ODEs parameterize the derivative of the hidden state with respect to time as a neural network:

\begin{equation}
\frac{d\mathbf{z}(t)}{dt} = \mathcal{F}_\theta(\mathbf{z}(t), t),
\end{equation}
where $\mathbf{z}(t)$ denotes the latent representation at time $t$ and $\mathcal{F}_\theta(\cdot)$ is a neural network modeling the derivative with respect to time. 

To provide physical intuition, the latent variable $\mathbf{z}(t)$ can be viewed as an abstract representation of an object’s physical state, implicitly encoding both position-like and velocity-like information:
$\mathbf{z}(t) = [\, \mathbf{x}(t),\, \mathbf{v}(t) \,]$,
where $\mathbf{x}(t)$ denotes the position and $\mathbf{v}(t)$ denotes the velocity of the object. A typical physical dynamical system can be formulated as:
\begin{equation}
\frac{d\mathbf{x}}{dt} = \mathbf{v},
\quad
\frac{d\mathbf{v}}{dt} = \mathbf{a} = f(\mathbf{x}, \mathbf{v}, t),
\end{equation}
where $\mathbf{a}$ represents the acceleration determined by a force field or other latent physical interactions.

Analogously, the Neural ODE function $\mathcal{F}_\theta$ learns to approximate this process within the latent space:
\begin{equation}
\mathcal{F}_\theta(\mathbf{z}(t), t)
\;\approx\;
\Big[\,
\mathbf{v}(t),\,
f_\theta\big(\mathbf{x}(t), \mathbf{v}(t), t\big)
\,\Big].
\end{equation}
allowing the model to capture both short-term velocity trends and long-term acceleration effects in a differentiable and physically consistent manner.

\paragraph{Motion Modeling and Prediction.}
After disentangling motion dynamics from visual appearance, an essential step is to model how objects evolve continuously over time.
Real-world motion is inherently continuous and physically constrained.
Frame-wise representations break temporal continuity and prevent the model from capturing latent quantities such as velocity and acceleration. To overcome this limitation, we incorporate Neural ODEs.

Specifically, given a sequence of motion features extracted from the video,
$\{\, {f}_{\text{mot}}^{(t)} \,\}_{t=1}^{T}$, we project them into a latent dynamical space through a learnable mapping network:
\begin{equation}
\mathbf{z}(t) = \Phi\psi\big(f_{\text{mot}}^{(t)}\big).
\end{equation}
where $\Phi\psi(\cdot)$ denotes a lightweight encoder implemented as a multi-layer perceptron (MLP) with LayerNorm and GELU activations. $\mathbf{z}(t)$ represents the latent motion state of the video at time t.

Subsequently, the Neural ODE module employs a learnable dynamical function $\mathcal{F}_\theta$ to describe the continuous temporal evolution of the latent motion state:
\begin{equation}
\mathbf{z}({t+\tau}) = \mathrm{ODESolve}\big(\mathcal{F}_\theta,\, \mathbf{z}(t),\, \tau\big),
\qquad \tau = 1, \dots, N,
\end{equation}
where $\mathrm{ODESolve}(\cdot)$ denotes a differentiable ODE solver.
This integration process can be interpreted as accumulating the latent dynamical field over the interval $[t,\, t+\tau]$, yielding a smooth, continuous, and differentiable latent trajectory.

Accurately modeling dynamic motion in videos requires more than passively encoding past observations; it demands the capacity to predict how objects will evolve under latent physical laws. For example, the future position of a moving object is highly sensitive to whether it follows constant velocity, uniform acceleration, or deviates due to external forces. As time progresses, the discrepancy between constant and accelerated motion trajectories becomes increasingly significant.
The ability to forecast future states under physical constraints serves as a strong indication that the model has captured the underlying dynamic motion.
To stabilize training, we apply gradient clipping to ODE parameters and we implement the function $\mathcal{F}_\theta$ following the standard design~\cite{chen2018neural}.

To map the latent trajectory back to the observable motion feature space, we introduce a lightweight readout network $R_\phi(\cdot)$ that produces the predicted motion feature sequence:
\begin{equation}
\hat{f}_{\text{mot}}^{(t+\tau)} = R\big(\mathbf{z}(t+\tau)\big),
\qquad \tau = 1, \dots, N.
\end{equation}

\paragraph{Self-supervised Training Strategy.}
To overcome the severe scarcity of physical labels in the data, our supervision strategy adopts an effective form of self-supervised learning. Accordingly, we supervise the ODE output trajectory $\{\hat{f}_{\text{mot}}^{(t+\tau)}\}_{t=1}^T$ by aligning it with the motion encoder outputs $\{f_{\text{mot}}^{(t+\tau)}\}_{t=1}^T$ using a mean squared error (MSE) loss:
\begin{equation}
\mathcal{L}_{\text{phys}} = \sum_{t=N}^{T-N} \sum_{\tau=1}^{N} \left\| f_{\text{mot}}^{(t+\tau)} - \hat{f}_{\text{mot}}^{(t+\tau)} \right\|^2
\label{ode}\end{equation}
It is worth noting that the loss is only computed over the valid prediction region. Specifically, we exclude the first $N$ frames, which do not have sufficient historical context for a full prediction window, and the last $N$ frames, where ground truth features are unavailable beyond time step $T$. As a result, the total physics loss is computed over the interval $t \in [N, T-N]$ with prediction horizon $\tau \in [1, N]$. This physics-consistent objective loss encourages the ODE module to learn latent physical dynamics that are consistent with the observed motion evolution. And the module is fully differentiable and jointly trained with the motion encoder in an end-to-end manner.

\subsection{Physics-aware Tokenization}
After obtaining disentangled motion and appearance features, denoted by $F_{\text{mot}}$ and $F_{\text{app}}$, we project them into the embedding space of an LLM. To achieve this, we introduce two specific linear projection heads:
\begin{equation}
Z^{\text{m}} = g_{\text{mot}}(F_{\text{mot}}), \quad Z^{\text{a}} = g_{\text{app}}(F_{\text{app}}), \quad 
\end{equation}
where $Z^{\text{m}} \in \mathbb{R}^{N \times d}$ and $Z^{\text{a}} \in \mathbb{R}^{N \times d}$ are two projected token sequences and $d$ is the hidden dimension of the LLM.

To guide PhyVLLM focus on each feature independently, we insert the projected features into the input stream at designated anchor positions using special tokens \texttt{<motion>} and \texttt{<appearance>}. These tokens are embedded within a prompt. For example, a representative prompt is:
\begin{quote}
\textit{“Appearance features: \texttt{<appearance>}, Motion features: \texttt{<motion>}. Can you describe this video?”}
\end{quote}

Formally, the final input to the LLM becomes:
\begin{equation}
\textit{input}\!=\![w_1, \!\cdots\!, w_{j-1}, Z^{\text{a}}, w_{j+1},\! \cdots\!, w_{k-1}, Z^{\text{m}}, w_k, \! \cdots\!, w_M]
\end{equation}
where $\{w_i\}_{i=1}^{M}$ are the token embeddings of the prompt, and $j, k$ indicate the insertion indices of the appearance and motion features. 

\subsection{Training}

\paragraph{Training Objectives.}
The training process is designed for physical modeling, motion-appearance disentanglement, and language understanding.
To ensure temporal coherence and physical motion modeling, we supervise the ODE-based motion prediction using a \textbf{physics-consistent loss} $\mathcal{L}_{\text{phys}}$, as defined in Eq.~\ref{ode}. To effectively disentangle motion cues from static visual appearance, we introduce an \textbf{appearance disentanglement loss} $\mathcal{L}_{\text{app}}$, as defined in Eq.~\ref{app_loss}.


To better align responses with instructions, we update the LLM via LoRA \cite{hu2022lora} with  $\mathcal{L}_{\text{LM}}$ (next-token cross-entropy). The overall objective is:
\begin{equation}
\mathcal{L}_{\text{total}} = \mathcal{L}_{\text{LM}} + \lambda\mathcal{L}_{\text{phys}} + \lambda\mathcal{L}_{\text{app}}
\end{equation}
where the $\lambda$ coefficients balance physical modeling, appearance disentanglement, and instruction tuning.


\paragraph{Instruction Tuning Datasets.}We construct a comprehensive instruction tuning dataset by integrating diverse sources from multiple video understanding tasks. Following InternVL2.5 \cite{chen2024far}, our dataset can be regarded as a subset of its video training corpus. Specifically, our dataset includes: Conversational video data collected from VideoChat \cite{li2023videochat} and Video-ChatGPT\cite{maaz2023video}; Video captioning data from Ego4D \cite{grauman2022ego4d} and YouCook2 \cite{das2013thousand}; Annotated video question answering samples from ActivityNet-QA~\cite{yu2019activitynet}, EgoQA~\cite{grauman2022ego4d}, and NextQA~\cite{xiao2021next}. In total, we utilize approximately 223k video instruction data samples to perform supervised fine-tuning (SFT) for our PhyVLLM model, which is about one-sixth the size of the dataset used in InternVL2.


\section{Experiments}

\begin{table*}[t]
\centering 
\caption{Comparisons with state-of-the-art methods on the PhyBench dataset.}
\footnotesize
\vspace{-0.2cm} 
\begin{tabular}{p{100pt}|>{\hfil}p{27pt}<{\hfil}|>{\hfil}p{35pt}<{\hfil}|>{\hfil}p{45pt}<{\hfil}>{\hfil}p{45pt}<{\hfil}cccc} 
\toprule
\textbf{Model} & \textbf{Size} & \textbf{Avg} & Accelerated & Decelerated & Uniform & Rebound & Parabolic \\
\midrule
GPT-4o & - & 34.05	& 0.87	& 0.11	& 10.3	& 63.92	& 95.03 \\
InternVL2.5 \cite{chen2024far}& 8B & 23.16 & 
0.00 & 0.00 & 99.67 & 1.00 & 0.00  \\
Qwen2.5-VL \cite{bai2025qwen2}& 7B & 22.90 & 0.44 & 0.78 & 73.00 & 34.50 & 3.00  \\
mPLUG-Owl3 \cite{ye2025mplugowl}& 7B & 19.72 &
1.33 & 0.00 & 63.22 & 31.33 & 0.00  \\
VideoChatGPT \cite{maaz2023video} & 7B & 17.47 & 7.89 & 6.90 & 4.00 & 65.50 & 19.83 \\
VideoLLaMA2 \cite{cheng2024videollama}  & 7B & 23.19 & 
0.11 & 0.00 & 99.22 & 1.17 & 0.50 \\
LLaVa-NeXT-Video \cite{zhang2024llavanextvideo} & 7B & 21.34 &
15.00 & 17.80 & 43.33 & 18.50 & 6.00  \\
InternVideo2 \cite{wang2024internvideo2} & 8B & 23.36 & 
1.11 & 1.33 & 95.67 & 4.67 & 0.00  \\
\rowcolor{blue!8} PhyVLLM & 7B & 40.52 & 48.67 & 45.49 & 15.11 & 82.83 & 16.67 \\
\midrule
InternVL2-finetune & 8B & 23.16 & 
0.00 & 0.00 & 99.67 & 1.00 & 0.00  \\
Qwen2.5-VL-finetune \cite{bai2025qwen2}& 7B & 46.72 & 18.89 & 20.11 & 69.00 & 78.33 & 63.33  \\
\rowcolor{blue!8} PhyVLLM-finetune & 7B & 79.33 & 77.67 & 68.97 & 81.33 & 92.00 & 81.67 \\
\bottomrule
\end{tabular}
\vspace{-0.2cm} 
\label{regular_performance_1}
\end{table*}

\subsection{PhyBench}
Despite the rapid progress of Video LLMs in various multimodal understanding tasks, there remains a fundamental gap in evaluating their physical dynamics reasoning capabilities. Existing benchmarks such as MVBench, NExT-QA, and Ego4D are primarily designed for appearance-centric tasks, focusing on object recognition, scene understanding, or high-level event description. Currently, there is no dedicated benchmark to assess the physical modeling capabilities of Video LLMs. This lack of targeted evaluation makes it difficult to quantify progress in physics-aware video understanding. To make matters worse, annotating physical quantities in real-world video datasets is extremely challenging. Attributes such as acceleration, velocity, or force are rarely directly observable from RGB videos, and collecting such data typically requires motion capture systems, sensors, or high-precision tracking methods, which do not scale effectively to large datasets.

To fill this gap, we propose PhyBench, a synthetic video-language benchmark specifically designed to evaluate the physical reasoning capabilities of Video LLMs. By leveraging a physics simulation platform\cite{kang2024far}, PhyBench enables fine-grained control over object motion, allowing us to systematically construct physical scenarios with accurate ground-truth dynamics. 
PhyBench selects relatively simple physical interaction scenarios to provide a controlled environment, while minimizing the influence of complex visual factors. The dataset covers five fundamental types of physical motion that serve as the basis of classical Newtonian mechanics. These include uniform motion, accelerated motion, decelerated motion, parabolic motion, and bouncing motion~\footnote{These five types of physical motion are predefined. The data generation platform can synthesize a broader range of physical motions (e.g., rotation, compound collisions), thereby providing more challenging reasoning environments.}.
Each video lasts approximately 2 to 60 seconds at a frame rate of 10 FPS and focuses on a single type of motion. The camera is fixed to eliminate view-based confounders, and objects are rendered with consistent lighting and material properties to ensure visual uniformity. 


\subsection{Other Benchmark} 
To comprehensively evaluate the performance of PhyVLLM, we also conduct experiments on several general video understanding benchmarks. \textbf{Video-MME}\cite{fu2025video} is a benchmark designed to evaluate the video analysis capabilities of MLLMs. It includes a diverse range of video types across various domains and durations. We report the results under the “without subtitle” setting. \textbf{MVBench}\cite{li2024mvbench} is an open-world video understanding benchmark aimed at assessing the temporal awareness of MLLMs. It covers 20 video tasks, ranging from perception to cognition, which cannot be effectively addressed using a single frame. 


\subsection{Implementations}
In our implementation, we adopt InternLM-7B~\cite{cai2024internlm2} as the base large language model. We adopt the fourth-order Runge–Kutta (RK4) as ODE solver. PhyVLLM is trained for 2 epochs on instruction tuning datasets using the AdamW optimizer with a learning rate of $2 \times 10^{-5}$ and a batch size of 2. For LoRA-based adaptation, we set the rank parameter to $r = 16$ and the scaling factor to $\alpha = 32$. During training, only the adapters and encoders are updated, while the backbone language model remains frozen. All experiments are conducted on 4 NVIDIA A800 GPUs.

\begin{table}[t]
\centering 
\caption{Comparisons with state-of-the-art methods on several general video understanding benchmarks.}
\footnotesize
\vspace{-0.2cm} 
\begin{tabular}{p{100pt}|>{\hfil}p{50pt}<{\hfil}>{\hfil}p{50pt}<{\hfil}} 
\toprule
\textbf{Model} & Video-MME & MVBench  \\
\midrule
\rowcolor{gray!8}\multicolumn{3}{c}{\textit{MLLMs}} \\
InternVL2.5 \cite{chen2024far}& 64.2 & 72.0   \\
Qwen2.5-VL \cite{bai2025qwen2}& 65.1 & 69.6  \\
mPLUG-Owl3 \cite{ye2025mplugowl}& 53.5 & 54.5   \\
InternVL3 \cite{chen2024far}& 66.3 & 75.4   \\
\midrule
\rowcolor{gray!8}\multicolumn{3}{c}{\textit{Video LLMs (16 frames input, except VideoChatGPT uses 100)}} \\
VideoChatGPT \cite{maaz2023video} & - & 32.7  \\
VideoLLaMA2 \cite{cheng2024videollama}  & 46.6 & 54.6  \\
LLaVa-NeXT-Video \cite{zhang2024llavanextvideo} & 35.6 & 46.0  \\
ST-LLM \cite{ST-LLM} & 37.9 & 54.9  \\
InternVideo2 \cite{wang2024internvideo2} & -  & 67.2   \\
VITA-1.5 \cite{fu2025vita} & 56.8 & 55.4  \\
\rowcolor{blue!8} PhyVLLM & 68.1 & 75.1  \\
\bottomrule
\end{tabular}
\vspace{-0.2cm} 
\label{regular_performance_2}
\end{table}

\subsection{Results on PhyBench}
All models are evaluated on the PhyBench. The task is formulated as a multiple-choice question answering problem, where accuracy is reported across five motion categories: acceleration, deceleration, uniform motion, bouncing, and parabolic motion. The results are summarized in Table~\ref{regular_performance_1}.

We first evaluate models under the zero-shot setting, without any task-specific fine-tuning, and compare our method with several MLLMs and video LLMs. The MLLMs perform poorly on this task. These models process each frame independently and lack explicit modeling of temporal or physical dynamics. Although InternVL2 achieves the highest average score among MLLMs (23.16), this is misleading: it nearly always predicts “uniform motion” for all samples, which inflates performance on that specific class but fails to generalize to other motion types.
In contrast, video LLMs perform slightly better overall, as they incorporate temporal information through multi-frame inputs. However, their performance on physical reasoning tasks remains limited. A major contributing factor is their tendency to rely on visual cues, without explicitly modeling the underlying motion dynamics. 
For instance, models such as VideoLLaMA2 and InternVideo2 often default to high-confidence predictions of uniform motion, failing to capture acceleration or deceleration. 
Our proposed PhyVLLM outperforms all baselines by a significant margin, achieving an average accuracy of $40.52$, with consistently strong performance across all five motion types. Notably, the model shows clear improvements on accelerated and decelerated motion, which are the most challenging types as they require understanding of second-order dynamics (i.e., changes in velocity over time). Furthermore, after fine-tuning, PhyVLLM continues to outperform two strong baselines, demonstrating its robustness and adaptability.  

\subsection{Results on General Video Understanding}
To comprehensively evaluate the performance of PhyVLLM, we also conduct experiments on several general video understanding benchmarks. The results on the Video-MME and MVBench are shown in Table~\ref{regular_performance_2}.
PhyVLLM adopts the same two-stage training pipeline as InternVL2.5 \cite{chen2024far} but is trained with only one-sixth of the instruction tuning data. Despite this drastic reduction, PhyVLLM achieves 68.1 on Video-MME and 75.1 on MVBench, outperforming most existing MLLMs and Video LLMs. 
We attribute this success to two key design choices: (1) the introduction of a Neural ODE module, which explicitly models motion as a continuous trajectory over time, and (2) the incorporation of a self-supervised training paradigm and disentangled motion-appearance representations, which guide the model toward learning structured physical priors. Together, these components enable PhyVLLM to go beyond frame-level cues and develop a deeper, physically grounded understanding of object motion while improving generalization across general video understanding tasks. 

\begin{table}[t]
    \centering
    \footnotesize
    \caption{The effectiveness of components.}\label{ab}
    \vspace{-0.2cm} 
    \begin{tabular}{p{128pt}>{\hfil}p{38pt}<{\hfil}>{\hfil}p{38pt}<{\hfil}}
    \toprule
        Method & PhyBench & MVBench\\ \midrule
        $base$ & 23.10 & 56.3 \\ 
        $base+L_{phys}$ & 69.32 & 64.5 \\ 
        $base+L_{app}$ & 43.37 & 67.3 \\   \rowcolor{blue!8}$base+L_{phys}+L_{app} (PhyVLLM)$ & 79.33 & 75.1 \\
        \bottomrule
    \end{tabular}
    \vspace{-0.2cm} 
\end{table}



\subsection{Ablations}
\paragraph{Effectiveness of each Component.}
As shown in Table~\ref{ab}, we perform a comprehensive ablation study to evaluate the contribution of each key component in our framework. The experiments on PhyBench are conducted after fine-tuning the model.
From the table, we can see that:
\begin{itemize}[leftmargin=1.2em]
\item The baseline model ($base$) adopts a minimal configuration, where a lightweight MLP is inserted between the visual encoder and the language model. This simple setup achieves an average score of $23.10$, serving as the reference point.

\item Introducing the physics-consistent loss ($base+L_{phys}$), which employs only a single encoder but integrates ODE-based modeling after the encoder. This strategy significantly improves performance, reaching $69.32$ on PhyBench and $64.5$ on MVBench, which confirms the importance of capturing temporal dynamics and physical consistency for video understanding. 

\item In parallel, we test the impact of appearance-motion disentanglement without incorporating ODE modeling ($base+L_{app}$). This configuration achieves $43.37$ on PhyBench and $67.3$ on MVBench, which highlights the advantage of structured feature separation.

\item Our full model PhyVLLM ($base+L_{phys}+L_{app}$) combines both physics-aware modeling and motion-appearance disentanglement in a unified manner. It achieves the highest performance of $79.33$ on PhyBench and $75.1$ on MVBench, demonstrating that the two components are complementary. ODE modeling contributes temporal coherence grounded in physical laws, while disentangled representations facilitate more interpretable and robust feature learning.
\end{itemize}

\paragraph{Effectiveness of Physical-guided Motion Prediction.}
To evaluate the effectiveness of Motion Prediction with Neural ODEs, we design the following experiment. We first feed the complete ground-truth frames T0–T11 of multiple videos into the motion encoder to extract their feature representations. For the same videos, we then use frames T0–T8 as input to predict motion feature representations for T9'–T11'. Finally, we compute the similarity between the predicted features and the ground-truth features, and report the average as the evaluation metric.

\begin{figure}[t]
\centering
\begin{minipage}{0.7\linewidth}\centering
\centerline{\includegraphics[height=2.4cm]{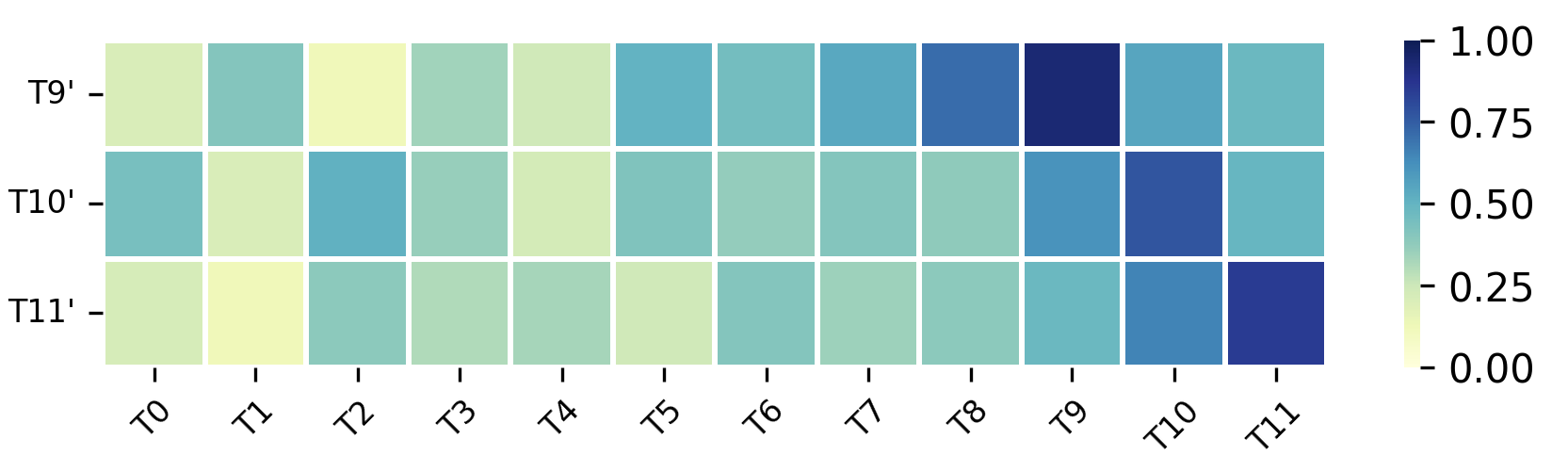}}
\end{minipage}
\vspace{-0.0cm} 
\caption{Similarity heatmap between predicted motion features \text{(T9'–T11')} and ground-truth motion features (T0–T11). Darker colors indicate higher similarity.}\vspace{-0.0cm} \label{similar}\medskip\end{figure}

The similarity heatmap is shown in Figure~\ref{similar}. It can be observed that the predicted frames T9'–T11' exhibit high similarity with the corresponding ground-truth frames, with the similarity concentrated along the diagonal and its neighboring regions. This indicates that under the self-supervised training mechanism, the physics-guided motion prediction effectively captures dynamic and accurately forecasts the object’s future physical states. 

\section{Conclusions}
In this paper, we propose PhyVLLM, a physics-guided video-language framework that enhances the physical reasoning capabilities of Video LLMs. To achieve this, we propose a motion-appearance disentanglement module that separates dynamic motion cues from static visual appearance, enabling the model to isolate physically meaningful features. To further capture the underlying physics, we incorporate a Neural ODE module to represent object motion as continuous-time trajectories. The resulting physics-aware representations are injected into a frozen LLM via lightweight adapter layers, ensuring efficient fine-tuning and smooth integration with the existing LLMs. To evaluate the proposed framework, we introduce PhyBench, a synthetic benchmark specifically designed to assess physical reasoning across canonical motion patterns under controlled conditions.
Extensive experiments on PhyVLLM and general video understanding benchmark validate the effectiveness of our approach, significantly improving the physical understanding capabilities of Video LLMs.
{
    \small
    \bibliographystyle{ieeenat_fullname}
    \bibliography{main}
}




\end{document}